\documentclass[journal]{IEEEtran}

\usepackage{cite}
\usepackage{amsmath,amssymb,amsfonts}
\usepackage{algorithm} 
\usepackage{graphicx}
\usepackage{xcolor}
\usepackage{colortbl}
\usepackage{ragged2e}
\colorlet{BLUE}{blue}
\usepackage{caption}
\usepackage{subcaption}
\usepackage{textcomp}
\usepackage{array}
\usepackage{textcomp}
\usepackage{stfloats}
\usepackage{flafter}
\usepackage{placeins}
\usepackage{url}
\usepackage{verbatim}
\usepackage{amsthm, bm, hyperref, mathrsfs, indentfirst, longtable}
\hypersetup{hidelinks}
\usepackage{svg}
\usepackage{longtable}
\usepackage{tabularx}
\usepackage{makecell}
\usepackage{booktabs,float}
\usepackage{bbm}
\usepackage{threeparttable}
\usepackage{multirow}
\usepackage{booktabs}
\usepackage{balance}

\begin{document}
\title{Vision--Wireless Fusion for Multi-User Localization: A Cross-Modal Transformer Approach}

\author{
    Can Zheng, 
    Jiguang He,~\IEEEmembership{Senior Member,~IEEE,}
    Guofa Cai,~\IEEEmembership{Senior Member,~IEEE,}\\
    Henk Wymeersch,~\IEEEmembership{Fellow,~IEEE,} 
    M\'erouane Debbah,~\IEEEmembership{Fellow,~IEEE} 
    \thanks{Can Zheng and Jiguang He are with the School of Computing and Information Technology, Great Bay University, Dongguan Key Laboratory for Intelligence and Information Technology, and Great Bay Institute for Advanced Study (GBIAS), Dongguan 523000, China (e-mail: zc331\_@korea.ac.kr, jiguang.he@gbu.edu.cn).}
    \thanks{Guofa Cai is with the School of Information Engineering, Guangdong University of Technology, Guangzhou 510006, China (e-mail: caiguofa2006@gdut.edu.cn).}
    \thanks{Henk Wymeersch is with the Department of Electrical Engineering, Chalmers University of Technology, Gothenburg, Sweden. (emails: henkw@chalmers.se).}
    \thanks{M\'erouane Debbah is with  the Research Institute for Digital Future, Khalifa University, 127788 Abu Dhabi, UAE (email: merouane.debbah@ku.ac.ae).}
}


\maketitle

\begin{abstract}
Accurate multi-user localization is challenging in complex urban environments, where wireless measurements can become ambiguous under noise, blockage, and multipath, {while visual observations provide complementary spatial context.} {This paper presents a vision--wireless fusion framework for multi-user localization using pilot-indexed channel state information (CSI).} {Orthogonal pilot indices preserve the identities of the communicating UEs in the CSI-token sequence and localization outputs.} {The model encodes each pilot-indexed CSI observation as a query token and uses cross-attention to retrieve user-specific information from spatial visual memory.} Self-attention among CSI tokens further captures inter-user interactions, while the resulting multimodal representations are used for user-wise localization. Experiments on different datasets show consistent improvements over model-based, CSI-only, and multimodal-fusion baselines. {Further experiments evaluate the model under different wireless and visual conditions.}
\end{abstract}


\section{Introduction}


    \IEEEPARstart{A}{{ccurate}} user localization is a fundamental capability for intelligent transportation systems, as spatial awareness directly affects traffic management, collision avoidance, and autonomous driving path planning \cite{v2iloc,6G_loc,5g_loc}. However, global navigation satellite systems (GNSS) often suffer from severe performance degradation in dense urban environments. In urban canyon scenarios, signal blockage by high-rise buildings, strong multipath reflections, and signal attenuation can lead to large positioning errors, making GNSS insufficient for safety-critical applications that require reliable and high-precision localization \cite{v2iaoa}.
    
    {Classical wireless localization methods estimate geometric parameters, such as angle-of-arrival (AoA) and time-of-arrival (ToA), from array measurements} \cite{TOA}. Representative techniques include multiple signal classification (MUSIC) and estimation of signal parameters via rotational invariant techniques (ESPRIT), which exploit signal and noise subspace structure or rotational invariance for super-resolution parameter estimation \cite{MUSIC,ESPRIT}. In multiple-input multiple-output orthogonal frequency-division multiplexing (MIMO-OFDM) systems, these ideas are often further connected to angle-delay or beamspace representations, where discrete Fourier transform (DFT) and inverse discrete Fourier transform (IDFT) operations along the antenna and subcarrier dimensions yield a sparse virtual-domain characterization of the channel \cite{DFT}. This motivates practical baselines such as three-dimensional (3D)-FFT peak extraction, which estimates dominant angular-delay features and then maps them to user coordinates.  In addition, traditional localization and tracking systems often employ probabilistic state-estimation tools to improve robustness against noisy measurements and model uncertainty. A classical example is Kalman filtering for recursive state tracking and estimate refinement \cite{Kalman}, while Bayesian localization methods maintain and update a posterior belief over candidate positions using Bayesian inference \cite{MC, BF, Bayesianfusion, Vehicular_pos}. Although such methods are physically interpretable and effective under favorable calibration and high signal-to-noise-ratio (SNR) conditions, their accuracy degrades significantly in low SNR, strong multipath, and non-line-of-sight (NLoS) environments, where the dominant path may be weak, corrupted, or no longer geometrically representative \cite{Sayed2005NetworkBased,Yin2013RobustTOA}.

    Another important direction is channel state information (CSI) fingerprinting, which replaces AoA and ToA parameter extraction with data-driven mapping from CSI observations to spatial coordinates \cite{FIFS, LiFS, Pilot, MonoPHY}. Representative methods such as DeepFi and PhaseFi learn location-dependent CSI fingerprints using deep neural networks (DNNs) \cite{DeepFi,PhaseFi}. Compared with classical signal processing methods, fingerprinting methods are more flexible in capturing nonlinear propagation-location relationships and {can alleviate some errors caused by simplified propagation models.} More recently, {spatial-context aware dynamic fusion network with mixture-of-experts (SCADF-MoE)} further extends this direction by introducing spatial-context-aware modeling and dynamic mixture-of-experts (MoE) fusion for frequency-robust wireless localization \cite{SCADF}. However, such methods still rely primarily on wireless propagation signatures and geometric side information, which may become highly ambiguous in complex urban scenes. Moreover, fingerprinting methods typically lack geometric context and often suffer from environment dependence, domain shift, and degraded robustness when propagation statistics change across scenes or blockage conditions \cite{DeepFi,PhaseFi}.

    Images provide geometric context and scene semantics that CSI alone cannot recover. Recent studies on vision-aided communications have shown that this context can improve physical-layer tasks \cite{ViWi,Camera,VisionPosition}. Our earlier work in \cite{WCNC_Loc} studied {vision--wireless fusion} for urban vehicle-to-infrastructure (V2I) localization through multimodal regression. Feature-level fusion of WiFi CSI and images has also been explored for indoor localization \cite{wifi}. Most existing methods, however, address single-user link-level prediction, beam selection, or localization. The matching-based framework in \cite{Huan2025TMC} addresses the missing one-to-one correspondence through image-assisted supervision, but its modal interaction occurs mainly between predicted coordinates and image-derived positions. The method in \cite{Huan2025TCOMM} instead converts image-derived azimuths into pretraining targets for semi-supervised CSI learning. In this design, vision supplies coarse directional supervision rather than instance-level correspondence between RF signals and visual objects. Both approaches assume that all communicating users are detectable in the image and that the detected objects correspond only to communicating users. {User-specific interaction between pilot-indexed CSI tokens and spatial visual features remains unexplored under these assumptions.}

    {We develop a vision--wireless fusion framework for multi-user localization using pilot-indexed CSI. Pilot-indexed CSI preserves communicating UE identity, while spatial visual memory provides scene context. Cross-attention uses each CSI token to retrieve visual information for the corresponding UE before predicting its 3D position. The model therefore performs user-wise localization rather than unordered visual-object matching.} The main contributions are as follows:
    
    \begin{itemize}
        \item We propose a {cross-modal Transformer-based vision--wireless fusion framework for multi-user localization using pilot-indexed CSI. Pilot-indexed CSI tokens} serve as queries to spatially encoded visual memory, {allowing the model to retrieve user-conditioned visual information for each CSI token.} Self-attention among CSI tokens further models inter-user interactions, while masking supports a variable number of active users.

        \item {We compare direct fusion with two feature alignment designs: a contrastive language--image pre-training (CLIP)-inspired alignment design and a modality-invariant and modality-specific representations for multimodal sentiment analysis (MISA) design \cite{CLIP,misa}. The CLIP-inspired design encourages paired representations to share a common embedding space, while MISA separates shared and modality-specific information. The experiments further show that the benefit of explicit alignment is dataset-dependent.}

        \item We conduct extensive evaluations on different datasets under different propagation and visual conditions. Beyond overall localization accuracy, we study CSI encoder selection, SNR and multipath robustness, visual dependence, active-user count, LoS/NLoS conditions, weather changes, and computational complexity. These results verify the contribution of both modalities and characterize when the proposed cross-modal design provides the largest gains.
    \end{itemize}

    The remainder of this paper is organized as follows. Section~\ref{II} introduces the system model and formulates the {multi-user localization problem using pilot-indexed CSI and visual observations}. Section~\ref{III} presents the proposed cross-modal localization architecture, including the vision encoder, CSI encoder, Transformer-based fusion module, localization head, and training strategy. Section~\ref{IV} reports the simulation settings and experimental results. Finally, Section~\ref{V} concludes this paper and discusses possible future research directions.

    \begin{figure}[t]
        \centering
        \captionsetup{font=footnotesize}
        \includegraphics[width=0.5\textwidth]{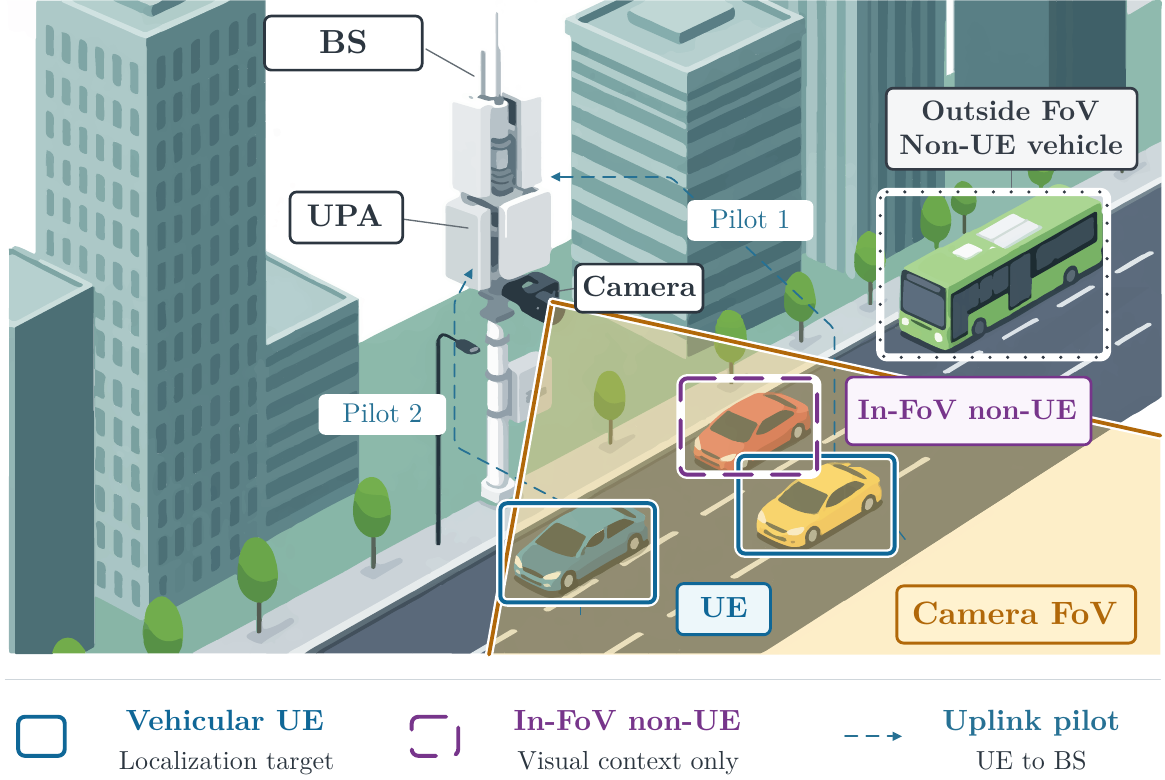}
        \caption{Multi-user V2I multimodal localization scenario. {All localization targets are vehicular UEs. Vehicles without communication links may also appear in the FoV.}}
        \label{fig:system_model}
    \end{figure}

    {\textit{Notations}: Unbolded italic symbols denote scalars, indices, functions, or parameter collections as specified in the text. Bold lowercase letters denote vectors, while bold uppercase letters denote matrices or higher-order tensors. Calligraphic letters denote sets, mappings, or loss functions as specified in the text. The sets of real and complex numbers are denoted by $\mathbb{R}$ and $\mathbb{C}$. The symbols $(\cdot)^\mathsf{T}$ and $(\cdot)^\mathsf{H}$ denote the transpose and Hermitian transpose. For real vectors $\mathbf{a}$ and $\mathbf{b}$, $\mathbf{a}^{\mathsf T}\mathbf{b}$ is their inner product. The operators $\Re\{\cdot\}$ and $\Im\{\cdot\}$ extract the real and imaginary parts of a complex quantity. The functions $\exp(\cdot)$, $\log(\cdot)$, $\sin(\cdot)$, and $\cos(\cdot)$ have their standard meanings. The notation $\lfloor\cdot\rfloor$ denotes the floor operation. The norms $\|\cdot\|_2$ and $\|\cdot\|_F$ denote the Euclidean and Frobenius norms. For a set $\mathcal{S}$, $|\mathcal{S}|$ denotes its cardinality. The indicator $\mathbb{I}(\cdot)$ equals one when its argument is true and zero otherwise. The notation $[\mathbf{a};\mathbf{b}]$ denotes concatenation along the feature dimension. A bar denotes $\ell_2$ normalization. For a feature matrix, this normalization is applied row by row.}

\section{System Model}
\label{II}

    \subsection{Scenario Description}
    Consider a multi-user V2I communication system deployed in an urban environment, as illustrated in Fig.~\ref{fig:system_model}. {A base station (BS)} is equipped with a uniform planar array (UPA) of $N_{\mathrm{ant}} = N_x \times N_y$ antenna elements and a co-located high-resolution RGB camera providing a field of view (FoV). Here, $N_x$ and $N_y$ are the numbers of antenna elements along the two UPA axes, and $N_{\mathrm{ant}}$ is the total number of antenna elements. The camera captures the visual representation of the physical scene, while the BS estimates the wireless CSI of the communication users.

    At any observation instant, {the BS} may serve a dynamic set of active users. {We localize active vehicular UEs within the camera FoV.} Let $\mathcal{K}=\{1,2,\dots,K\}$ denote this in-FoV user set, where $K$ is the number of active in-FoV users. The number of users satisfies $1\leq K\leq K_{\max}$, where $K_{\max}$ is the maximum supported number of users. The position of user $k\in\mathcal{K}$ is $\mathbf{p}_k =[x_k,y_k,z_k]^\mathsf{T}\in \mathbb{R}^3$ in the BS-local coordinate system {with the BS as its origin}, where $x_k$, $y_k$, and $z_k$ are its three coordinate components.

    \begin{figure*}[t]
        \centering
        \captionsetup{font=footnotesize}
        \includegraphics[width=\textwidth]{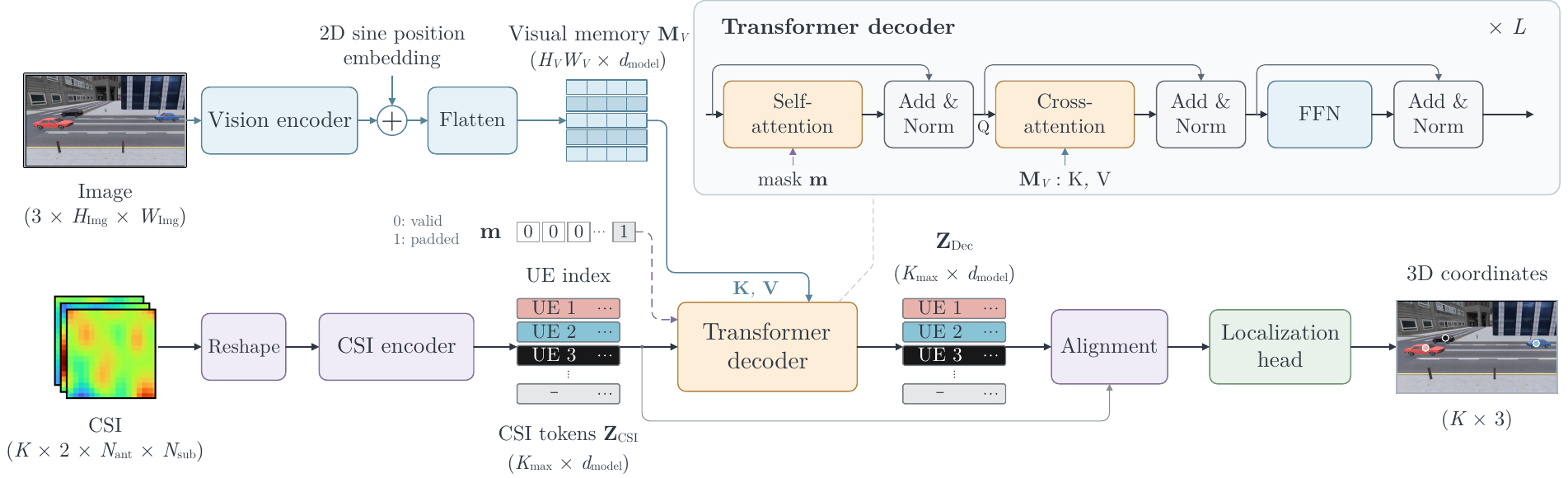}
        \caption{Architecture of the proposed multi-user vision--wireless localization model.}
        \label{fig:structure}
    \end{figure*}

    \subsection{Multi-User Observation Model}
    The uplink (UL) transmission employs OFDM over $N_{\mathrm{sub}}$ active subcarriers. To avoid multi-user interference during channel estimation, {the BS} allocates orthogonal UL resources to different active users, utilizing orthogonal sounding reference signals (SRSs) or distinct time-frequency resource blocks. {The pilot index identifies each UE and associates its CSI token with the corresponding localization output.}
    \begingroup
    
    For user $k\in\mathcal{K}$, the UL signal propagates through ${P_k}$ resolvable paths. Let $m\in\{1,\ldots,N_{\mathrm{ant}}\}$ index the antenna elements, $n\in\{0,\ldots,N_{\mathrm{sub}}-1\}$ index the subcarriers, and $\ell\in\{1,\ldots,{P_k}\}$ index the propagation paths. The complex baseband channel response from user $k$ to antenna element $m$ on subcarrier $n$ is modeled as
    \begin{equation}
        \tilde{h}_{k,m,n}
        =
        \sum_{\ell=1}^{{P_k}}
        \alpha_{k,\ell}
        e^{-j2\pi(f_c+n\Delta f)\tau_{k,\ell}}
        a_m(\theta_{k,\ell},\phi_{k,\ell}),
    \end{equation}
    where $f_c$ is the carrier frequency, and $\Delta f$ is the subcarrier spacing. {The variable $\alpha_{k,\ell}\in\mathbb{C}$ is the complex gain of path $\ell$, while $\tau_{k,\ell}$ denotes its time of arrival.} $a_m(\theta_{k,\ell},\phi_{k,\ell})$ is the $m$-th element of the UPA steering vector for elevation angle $\theta_{k,\ell}$ and azimuth angle $\phi_{k,\ell}$. {We assume ideal synchronization between each UE and the BS, so $\tau_{k,\ell}$ contains only the propagation delay and does not include clock bias.} Stacking the responses over all antennas and subcarriers gives the $k$-th user's complex CSI matrix $\tilde{\mathbf{H}}_k \in \mathbb{C}^{N_{\mathrm{ant}} \times N_{\mathrm{sub}}}$, whose $(m,n)$-th entry is $\tilde{h}_{k,m,n}$.
    \endgroup

    Synchronously with the wireless CSI acquisition, the co-located camera captures an RGB image of the surrounding road scene. This image is denoted by $\mathbf{X}_{\mathrm{Img}} \in \mathbb{R}^{3 \times H_{\mathrm{Img}} \times W_{\mathrm{Img}}}$, where $H_{\mathrm{Img}}$ and $W_{\mathrm{Img}}$ represent the image height and width, respectively.

    The wireless and visual modalities provide complementary information for localization. Specifically, the CSI contains user-specific propagation characteristics over the spatial and frequency domains, while the image provides comprehensive geometric and semantic context regarding the physical environment. Nevertheless, wireless signals may become ambiguous under severe NLoS and rich scattering conditions, and visual observations alone cannot explicitly reveal the communication identity of individual vehicular users.

    \subsection{Problem Formulation}

    Given the RGB image $\mathbf{X}_{\mathrm{Img}}$ and the CSI observations of all active vehicular users, denoted by $\{\tilde{\mathbf{H}}_k\}_{k=1}^{K}$, the objective is to estimate the 3D positions of these active vehicular users. Formally, the multimodal localization framework aims to learn a mapping function
    \begin{equation}
        \{\hat{\mathbf{p}}_k\}_{k=1}^{K}
        =
        f_{\Theta}
        \left(
        \mathbf{X}_{\mathrm{Img}}, \{\tilde{\mathbf{H}}_k\}_{k=1}^{K}
        \right),
    \end{equation}
    where $f_{\Theta}(\cdot)$ represents a parameterized mapping function defined by a set of trainable parameters $\Theta$, and $\hat{\mathbf{p}}_k \in \mathbb{R}^3$ denotes the predicted position of the $k$-th vehicular user.

\section{Cross-Modal Localization Architecture}
\label{III}

    {{As illustrated in Fig.~\ref{fig:structure}, the proposed vision--wireless fusion framework performs multi-user localization using pilot-indexed CSI.} For the visual modality, the vision encoder extracts a spatial feature map from the RGB image. {The CSI encoder maps each pilot-indexed CSI observation to a query token. The Transformer decoder uses cross-attention to retrieve visual information for each UE from the spatial visual memory while preserving the user order.} {We compare direct concatenation with CLIP- and MISA-based alignment. A shared localization head maps each selected feature to a 3D position.}}

    \subsection{Vision Encoder}
    Given the input image $\mathbf{X}_{\mathrm{Img}} \in \mathbb{R}^{3 \times H_{\mathrm{Img}} \times W_{\mathrm{Img}}}$, a truncated ResNet-18 backbone is employed to extract visual features from the road scene \cite{ResNet}. This backbone encodes the raw image into a latent feature map
    $\mathbf{F}_V \in \mathbb{R}^{d_{\mathrm{model}} \times H_V \times W_V}$, where $d_{\mathrm{model}}$ denotes the hidden feature dimension, and $H_V$ and $W_V$ represent the spatial dimensions of the output feature map.
    
    \begingroup
    
    To preserve the spatial structure after flattening, a deterministic two-dimensional (2D) sinusoidal positional embedding (PE) is introduced. For feature-map coordinates $h\in\{1,\ldots,H_V\}$ and $w\in\{1,\ldots,W_V\}$, define the normalized coordinates $u_h=2\pi h/H_V$ and $u_w=2\pi w/W_V$. The vertical and horizontal PE components each contain $d_{\mathrm{model}}/2$ channels. For $u\in\{u_h,u_w\}$ and channel index $c\in\{0,\ldots,d_{\mathrm{model}}/2-1\}$, the component is
    \begin{equation}
        \mathrm{PE}(u,c)=
        \begin{cases}
        \sin\!\left(u/T_\mathrm{PE}^{2\lfloor c/2\rfloor/(d_{\mathrm{model}}/2)}\right), & c\text{ even},\\
        \cos\!\left(u/T_\mathrm{PE}^{2\lfloor c/2\rfloor/(d_{\mathrm{model}}/2)}\right), & c\text{ odd},
        \end{cases}
    \end{equation}
    where $T_\mathrm{PE}$ is the temperature constant. Concatenating the vertical and horizontal components forms the PE tensor.
    \endgroup
    By superimposing the PE onto the visual feature map, we obtain the position-enhanced visual representation
    \begin{equation}
        \mathbf F_V^{\mathrm{PE}}=\mathbf F_V+\mathbf P_V,
    \end{equation}
    where $\mathbf{P}_V \in \mathbb{R}^{d_{\mathrm{model}} \times H_V \times W_V}$ denotes the PE tensor. The resulting tensor is then flattened along the spatial dimensions to construct the visual memory sequence $\mathbf{M}_V\in\mathbb{R}^{H_V W_V \times d_{\mathrm{model}}}$.
    {The positional encoding retains spatial information after flattening.}

    \subsection{Wireless CSI Representation and Encoder}

    For the $k$-th user, the input CSI is given by the complex matrix
    $\tilde{\mathbf{H}}_k \in \mathbb{C}^{N_{\mathrm{ant}} \times N_{\mathrm{sub}}}$.
    To enable real-valued processing, the CSI is decomposed into its real and imaginary components and reconstructed as
    \begin{equation}
        \mathbf{H}_k =
        \begin{bmatrix}
            \Re\{\tilde{\mathbf{H}}_k\} \\
            \Im\{\tilde{\mathbf{H}}_k\}
        \end{bmatrix}
        \in \mathbb{R}^{2 \times N_{\mathrm{ant}} \times N_{\mathrm{sub}}}.
    \end{equation}
    
    {The CSI feature extractor maps the two-channel representation to a user-specific token through the CSI encoder $f_{\mathrm{CSI}}(\cdot)$}
    \begin{equation}
        \mathbf{z}_{\mathrm{CSI},k}=f_{\mathrm{CSI}}(\mathbf{H}_k)\in\mathbb{R}^{d_{\mathrm{model}}}.
    \end{equation}
    By stacking all user-wise tokens and zero-padding invalid slots when necessary, we obtain the CSI token sequence $\mathbf{Z}_{\mathrm{CSI}}=[\mathbf{z}_{\mathrm{CSI},1},\mathbf{z}_{\mathrm{CSI},2},\ldots,\mathbf{z}_{\mathrm{CSI},K_{\max}}]^\mathsf{T}\in\mathbb{R}^{K_{\max} \times d_{\mathrm{model}}}$.

    {To distinguish active users from padded slots, we define a binary padding mask $\mathbf{m}\in\{0,1\}^{K_{\max}}$ where $m_k=0$ indicates a valid user slot and $m_k=1$ denotes a padded slot. The padding mask excludes unused CSI slots from the keys and values in self-attention. Outputs at these slots are ignored when computing losses and localization errors.}This token sequence is directly fed into the Transformer decoder as the target input. In this way, each token preserves the user-specific wireless information extracted from CSI observations and serves as an instance-level query for the subsequent cross-modal fusion.

    \newcommand{\experimentalsettingstable}{%
    \begin{table}[t]

    \centering
    \caption{Experimental settings for all datasets.}
    \label{tab:experimental_settings}
    \scriptsize
    \renewcommand{\arraystretch}{1.02}
    \setlength{\tabcolsep}{2.5pt}
    \begin{threeparttable}
    \begin{tabularx}{\columnwidth}{@{}>{\raggedright\arraybackslash}p{0.15\columnwidth}>{\raggedright\arraybackslash}p{0.48\columnwidth}>{\raggedright\arraybackslash}X@{}}
    \toprule
    \textbf{Group} & \textbf{Parameter} & \textbf{DV6G / RMT / MW} \\
    \midrule
    \multirow{7}{*}{Data}
    & Carrier frequency & 60 / 60 / 28~GHz \\
    & BS array & $4\times4$ UPA \\
    & Subcarriers & 8 in 512 \\
    & \# paths & 10 \\
    & Image size & $3\times270\times480$ \\
    & $K_{\max}$ & 10 \\
    & $T_\mathrm{PE}$ & 10000 \\
    \midrule
    \multirow{9}{*}{Training}
    & Batch size & 16 \\
    & Epochs & 100 \\
    & CSI learning rate & $10^{-3}$ \\
    & Visual learning rate & $10^{-5}$ \\
    & Other learning rate & $10^{-4}$ \\
    & Weight decay & 0.01 \\
    & Model dimension $d_{\mathrm{model}}$ & 256 \\
    & Decoder layers/heads/FFN width & 3/4/1024 \\
    & Dropout & 0.1 \\
    \midrule
    \multirow{6}{*}{Alignment}
    & Projection dimension $d_{\mathrm{proj}}$ (CLIP/MISA) & 256/128 \\
    & Initial similarity log-scale $\gamma$ & $\log(1/0.07)$ \\
    & $\lambda_{\mathrm{CLIP}}$ & 0.02 \\
    & $\lambda_{\mathrm{sim}}$ & 0.02 \\
    & $\lambda_{\mathrm{diff}}$ & 0.05 \\
    & $\lambda_{\mathrm{rec}}$ & 0.10 \\
    \bottomrule
    \end{tabularx}
    \begin{tablenotes}[flushleft]
    \scriptsize
    \item \textit{Note:} DV6G, RMT, and MW denote DeepVerse 6G, Raymobtime, and Multimodal-Wireless datasets, respectively. {Dataset-specific values are listed in this order, separated by slashes. General settings are shared by all three datasets.}
    \end{tablenotes}
    \end{threeparttable}
    \end{table}
    }

    \newcommand{\architecturetable}{%
    \begin{table*}[t]

    \centering
    \caption{Layer configurations of all compared methods.}
    \label{tab:model_architectures}
    \scriptsize
    \renewcommand{\arraystretch}{1.05}
    \setlength{\tabcolsep}{4.0pt}
    \begin{tabularx}{\textwidth}{@{}>{\raggedright\arraybackslash}p{0.12\textwidth}>{\raggedright\arraybackslash}p{0.13\textwidth}>{\RaggedRight\arraybackslash}X>{\raggedright\arraybackslash}p{0.09\textwidth}@{}}
    \toprule
    \textbf{{Methods}} & \textbf{{Modalities}} & \textbf{Layer sequence} & \textbf{Output size} \\
    \midrule
    3D-FFT
    & \mbox{{CSI}}
    & \leavevmode{CSI $\rightarrow\mathrm{IFFT}_{2048}$ with zero padding $\rightarrow\mathrm{FFT2D}_{64\times64}$ over unit-disk direction candidates $\rightarrow$ joint angle--delay peak $\rightarrow$ 3D mapping.}
    & $3$ \\
    \arrayrulecolor{gray!45}\specialrule{0.25pt}{0.25mm}{0.25mm}\arrayrulecolor{black}
    MUSIC
    & \mbox{{CSI}}
    & \leavevmode{CSI $\rightarrow$ $16\times16$ spatial covariance and $8\times8$ frequency covariance $\rightarrow$ Eigenvalue decomposition (EVD) $\rightarrow$ direction spectrum and range spectrum $\rightarrow$ largest spatial and range peaks paired $\rightarrow$ 3D mapping.}
    & $3$ \\
    \arrayrulecolor{gray!45}\specialrule{0.25pt}{0.25mm}{0.25mm}\arrayrulecolor{black}
    MLP
    & \mbox{{CSI}}
    & Flatten $\rightarrow[\mathrm{FC}(256,512)+\mathrm{LN}+\mathrm{GELU}+\mathrm{Drop}(0.1)]\times2\rightarrow\mathrm{FC}(512,256)\rightarrow$ CSI-Reg
    & $3$ \\
    \arrayrulecolor{gray!45}\specialrule{0.25pt}{0.25mm}{0.25mm}\arrayrulecolor{black}
    CNN
    & \mbox{{CSI}}
    & $[\mathrm{C2D}(2,32,3)\rightarrow\mathrm{C2D}(32,64,3)\rightarrow\mathrm{C2D}(64,128,3)]+\mathrm{BN}+\mathrm{GELU}\rightarrow\mathrm{AAP}(2,2)\rightarrow\mathrm{FC}(512,256)+\mathrm{LN}+\mathrm{GELU}\rightarrow$ CSI-Reg
    & $3$ \\
    \arrayrulecolor{gray!45}\specialrule{0.25pt}{0.25mm}{0.25mm}\arrayrulecolor{black}
    FNN
    & \mbox{{CSI}}
    & $\{\mathrm{C2D}(2,32,k)\rightarrow\mathrm{C2D}(32,64,k)+\mathrm{BN}+\mathrm{GELU}+\mathrm{GAP}\}_{k=(3,1),(1,3)}\rightarrow[\mathrm{Feat}_{3\times1};\mathrm{Feat}_{1\times3}]\rightarrow\mathrm{FC}(128,256)+\mathrm{LN}+\mathrm{GELU}\rightarrow$ CSI-Reg
    & $3$ \\
    \arrayrulecolor{gray!45}\specialrule{0.25pt}{0.25mm}{0.25mm}\arrayrulecolor{black}
    L2L
    & \mbox{{CSI}}
    & angle-delay power $\rightarrow$ horizontal/vertical C3D refinement $\rightarrow\mathrm{C3D}(64,128,3)\rightarrow\mathrm{Inc3D}(64,128,256)\rightarrow\mathrm{GAP}\rightarrow\mathrm{FC}(256,256)+\mathrm{LN}+\mathrm{GELU}\rightarrow$ CSI-Reg
    & $3$ \\
    \arrayrulecolor{gray!45}\specialrule{0.25pt}{0.25mm}{0.25mm}\arrayrulecolor{black}
    VAR
    & \mbox{{Vision + CSI}}
    & ResNet-18 through layer3 $\rightarrow\mathrm{C2D}(256,256,1)+\mathrm{GAP}$, CSI encoder$^{\dagger}\rightarrow[\mathrm{GAP}(\mathbf{F}_V);\mathbf{z}_{\mathrm{CSI}}]\rightarrow\mathrm{Reg}(512)$
    & $K_{\max}\times3$ \\
    \arrayrulecolor{gray!45}\specialrule{0.25pt}{0.25mm}{0.25mm}\arrayrulecolor{black}
    Proposed (None)
    & \mbox{{Vision + CSI}}
    & Trunk $\rightarrow[\mathbf{Z}_{\mathrm{Dec}};\mathbf{Z}_{\mathrm{CSI}}]\rightarrow\mathrm{Reg}(512)$
    & $K_{\max}\times3$ \\
    \arrayrulecolor{gray!45}\specialrule{0.25pt}{0.25mm}{0.25mm}\arrayrulecolor{black}
    Proposed (CLIP)
    & \mbox{{Vision + CSI}}
    & Trunk $\rightarrow\{\mathrm{FC}(256,256)+\mathrm{ReLU}+\mathrm{FC}(256,256)+\ell_2\text{ norm}\}_{\mu\in\{\mathrm{Dec},\mathrm{CSI}\}}\rightarrow[\bar{\mathbf{U}}_{\mathrm{Dec}};\bar{\mathbf{U}}_{\mathrm{CSI}}]\rightarrow\mathrm{Reg}(512)$
    & $K_{\max}\times3$ \\
    \arrayrulecolor{gray!45}\specialrule{0.25pt}{0.25mm}{0.25mm}\arrayrulecolor{black}
    Proposed (MISA)
    & \mbox{{Vision + CSI}}
    & Trunk $\rightarrow$ two $\mathrm{FC}(256,128)+\mathrm{LN}+\mathrm{GELU}$ projections $\rightarrow$ shared/private $\mathrm{FC}(128,128)+\mathrm{GELU}$ encoders $\rightarrow[\mathbf{S}_{\mathrm{Dec}};\mathbf{C}_{\mathrm{Dec}};\mathbf{S}_{\mathrm{CSI}};\mathbf{C}_{\mathrm{CSI}}]\rightarrow\mathrm{Reg}(512)$, with two reconstruction heads used only by $\mathcal{L}_{\mathrm{rec}}$
    & $K_{\max}\times3$ \\
    \bottomrule
    \multicolumn{4}{@{}l}{\parbox{0.99\textwidth}{\vspace{0.4mm}\textit{Note:} Trunk: ResNet-18 through layer3 $\rightarrow\mathrm{C2D}(256,256,1)$ + 2D sine PE, CSI encoder$^{\dagger}$, and three decoder layers with $d_{\mathrm{model}}=256$, four heads, and an FFN width of 1024. Reg$(d)$: $\mathrm{FC}(d,128)\rightarrow\mathrm{FC}(128,64)\rightarrow\mathrm{FC}(64,3)$ with ReLU and dropout 0.1. CSI-Reg: $\mathrm{FC}(256,64)\rightarrow\mathrm{FC}(64,3)$ with GELU. Multimodal models produce $K_{\max}$ rows internally, while only the $K$ valid rows contribute to training and evaluation. $^{\dagger}$The selected main model uses CNN on DeepVerse 6G and FNN on Raymobtime and Multimodal-Wireless. All Raymobtime alignment methods in Table~\ref{tab:overall_three_datasets} use FNN.}} \\
    \end{tabularx}
    \end{table*}
    }

    \subsection{Cross-Modal Transformer Decoder}

    The proposed cross-modal fusion module is built upon a Transformer decoder with $L$ stacked layers. The decoder takes the CSI token sequence $\mathbf{Z}_{\mathrm{CSI}}$ as the target input and the visual memory $\mathbf{M}_V$ as the source memory. Since the CSI tokens are user-specific, they naturally serve as instance-level queries for retrieving the visual evidence relevant to each user.
    
    To describe the stacked decoder structure, let $\mathbf{Z}^{(0)}=\mathbf{Z}_{\mathrm{CSI}}$ denote the input token sequence to the first decoder layer. For the $l$-th decoder layer, where $l\in\{1,\dots,L\}$, the input token sequence is denoted by $\mathbf{Z}^{(l-1)}$. The multi-head self-attention (MHSA) sublayer first models the interactions among user tokens:
    \begin{equation}
        \mathbf{Z}_{\mathrm{SA}}^{(l)}
        =
        \mathrm{MHSA}\!\left(
        \mathbf{Z}^{(l-1)},
        \mathbf{Z}^{(l-1)},
        \mathbf{Z}^{(l-1)}
        \right).
    \end{equation}
    This step allows the model to capture inter-user correlations and refine the per-user wireless representations under multi-user scenarios.
    
    Next, the multi-head cross-attention (MHCA) sublayer uses the refined token sequence to query the visual memory:
    \begin{equation}
        \mathbf{Z}_{\mathrm{CA}}^{(l)}
        =
        \mathrm{MHCA}\!\left(
        \mathbf{Z}_{\mathrm{SA}}^{(l)},
        \mathbf{M}_V,
        \mathbf{M}_V
        \right).
    \end{equation}
    Through this cross-modal interaction, each CSI-derived token selectively attends to the most relevant spatial regions in the visual feature map and gathers scene-aware information for localization.
 
    \begin{figure*}[t]
        
        \captionsetup{font=footnotesize}
        \centering
        \begin{subfigure}[b]{0.485\textwidth}
            \centering
            \includegraphics[width=\linewidth]{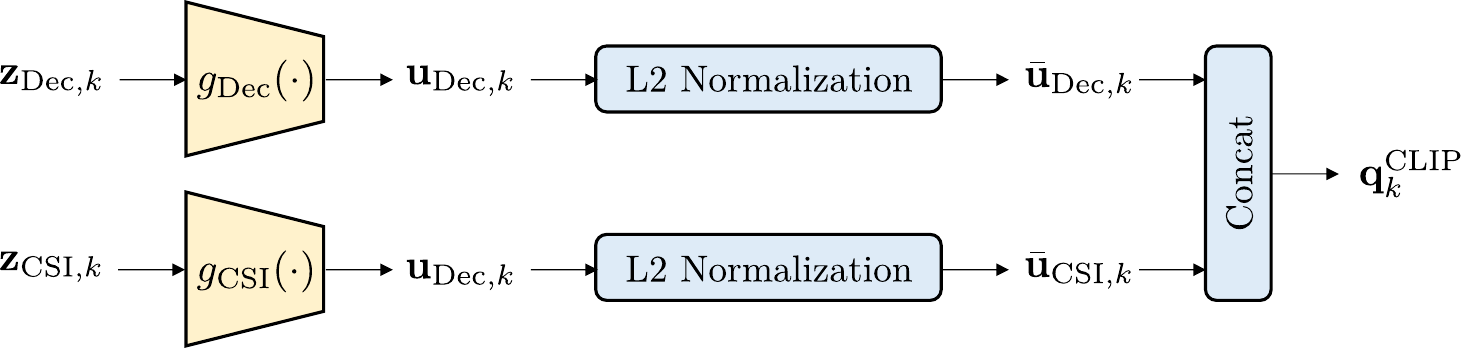}
            \caption{CLIP alignment}
            \label{fig:clip_alignment}
        \end{subfigure}\hfill
        \begin{subfigure}[b]{0.485\textwidth}
            \centering
            \includegraphics[width=\linewidth]{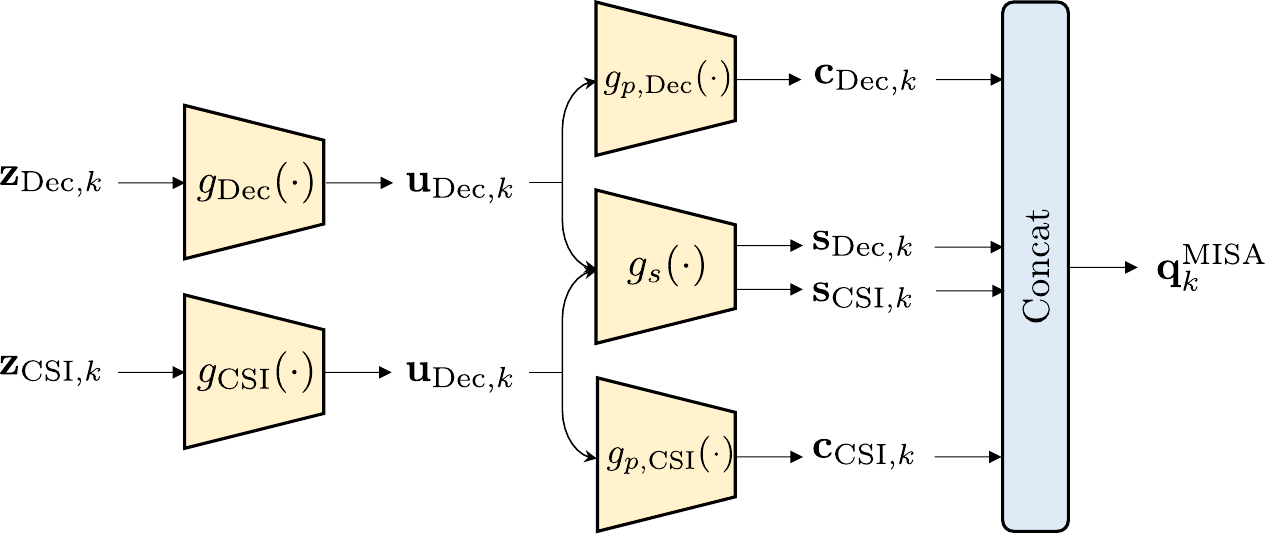}
            \caption{MISA alignment}
            \label{fig:misa_alignment}
        \end{subfigure}
        \caption{Cross-modal alignment methods in the proposed model.}
        \label{fig:alignment_variants}
    \end{figure*}
   
    The output of the $l$-th decoder layer is then obtained by applying a feed-forward network (FFN) to $\mathbf{Z}_{\mathrm{CA}}^{(l)}$, i.e.,
    \begin{equation}
        \mathbf{Z}^{(l)}
        =
        \mathrm{FFN}\!\left(
        \mathbf{Z}_{\mathrm{CA}}^{(l)}
        \right),
    \end{equation}
    {where the standard residual connections, dropout, and layer normalization in each decoder layer are omitted from the equations for brevity but are included in the implementation following the standard transformer decoder design {\cite{vaswani2017attention}}.} {Attention updates each token without changing its sequence position. Hence, output token $k$ remains associated with CSI token $k$ and the corresponding pilot-indexed UE.}

    After passing through all $L$ decoder layers, the final decoder output is given by
    \begin{equation}
        \mathbf{Z}_{\mathrm{Dec}}
        =
        \mathbf{Z}^{(L)}
        \in
        \mathbb{R}^{K_{\max}\times d_{\mathrm{model}}}.
    \end{equation}
    The resulting decoder tokens encode both the user-specific wireless characteristics inherited from CSI and the scene-aware visual context retrieved from the image.
    {For user $k$, the decoder token $\mathbf{z}_{\mathrm{Dec},k}$ and the corresponding CSI token $\mathbf{z}_{\mathrm{CSI},k}$ are passed to the Alignment block in Fig.~\ref{fig:structure}. Without alignment, the two tokens are directly concatenated and passed to the localization head.}

    \subsection{\texorpdfstring{{Alternative Alignment Designs}}{Alternative Alignment Designs}}

    \begingroup
    
    {{Besides directly concatenating the fused token and the CSI token for localization, we consider two alternative alignment schemes based on CLIP and MISA. } Let $\mathcal{M}=\{\mathrm{Dec},\mathrm{CSI}\}$ denote the two modality labels and let $\mu\in\mathcal{M}$ index a modality. For user $k$, a modality-specific head $g_\mu$ projects token $\mathbf{z}_{\mu,k}$ to $\mathbf{u}_{\mu,k}=g_\mu(\mathbf{z}_{\mu,k})\in\mathbb{R}^{d_{\mathrm{proj}}}$. Here, $\mathbf{z}_{\mathrm{Dec},k}$ is the fused decoder token and $\mathbf{z}_{\mathrm{CSI},k}$ is the original CSI token. The label $\mathrm{Dec}$ denotes the {decoded vision--wireless fusion representation}, while $\mathrm{CSI}$ denotes the {CSI representation before fusion}. Let $\mathcal{I}$ denote the indices of all valid tokens in the mini-batch. Stacking $\mathbf{u}_{\mu,i}^{\mathsf T}$ as rows gives $\mathbf{U}_\mu\in\mathbb{R}^{|\mathcal{I}|\times d_{\mathrm{proj}}}$. {The resulting CLIP or MISA feature is then passed to the localization head} \cite{CLIP, misa}. Fig.~\ref{fig:alignment_variants} shows the feature flow of both methods.}

    \subsubsection{CLIP Alignment}
    {CLIP treats the decoded multimodal representation and its paired CSI representation as two observations of the same user location. It maps them into a common embedding space so that representations associated with the same location are close, while unmatched user representations are separated. CLIP uses two independently parameterized projection heads and normalizes each projected vector as}
    \begin{equation}
        \bar{\mathbf{u}}_{\mu,k}
        =\frac{\mathbf{u}_{\mu,k}}{\|\mathbf{u}_{\mu,k}\|_2}.
    \end{equation}
    {The two normalized modal embeddings are concatenated to form the CLIP feature}
    \begin{equation}
        \mathbf{q}_{k}^{\mathrm{CLIP}}
        =\big[
        \bar{\mathbf{u}}_{\mathrm{Dec},k};
        \bar{\mathbf{u}}_{\mathrm{CSI},k}\big]
        \in\mathbb{R}^{2d_{\mathrm{proj}}}.
    \end{equation}
    {The same normalized embeddings are also used to calculate the CLIP alignment loss defined below.}

    \subsubsection{MISA Alignment}
    {MISA aligns information shared by the modalities but does not require all useful features to be common to both modalities. It also models modality-specific diversity. Accordingly, MISA uses a shared encoder $g_s(\cdot)$ and modality-specific private encoders $g_{p,\mu}(\cdot)$ to map each projected token to shared and modality-specific representations, respectively:}
    \begin{align}
        \mathbf{s}_{\mu,k}&=g_s(\mathbf{u}_{\mu,k}), \\
        \mathbf{c}_{\mu,k}&=g_{p,\mu}(\mathbf{u}_{\mu,k}),
    \end{align}
    {where $\mathbf{s}_{\mu,k},\mathbf{c}_{\mu,k}\in\mathbb{R}^{d_{\mathrm{proj}}}$. The vector $\mathbf{s}_{\mu,k}$ represents information shared by both modalities. The vector $\mathbf{c}_{\mu,k}$ represents complementary information specific to modality $\mu$.}

    {MISA concatenates the four shared and private representations in decoder and CSI order to form its localization feature}
    \begin{align}
        \mathbf{q}_{k}^{\mathrm{MISA}}
        =\big[\mathbf{s}_{\mathrm{Dec},k};\mathbf{c}_{\mathrm{Dec},k};
        \mathbf{s}_{\mathrm{CSI},k};\mathbf{c}_{\mathrm{CSI},k}\big]
        \in\mathbb{R}^{4d_{\mathrm{proj}}}.
    \end{align}

    \subsubsection{Alignment Losses}
    {Both methods use the following symmetric noise-contrastive estimation (NCE) loss \cite{gutmann2010noise, oord2018representation}. Consider two normalized feature matrices $\mathbf{A},\mathbf{B}\in\mathbb{R}^{|\mathcal{I}|\times d_{\mathrm{proj}}}$, where $i,j\in\mathcal{I}$ identify matrix rows. The NCE loss is}
    \begin{equation}
    \begin{split}
        \mathcal{L}_{\mathrm{NCE}}(\mathbf{A},\mathbf{B})
        =-\frac{1}{2|\mathcal{I}|}\sum_{i\in\mathcal{I}}
        \bigg(&\log\frac{\exp(s_{ii})}{\sum_{j\in\mathcal{I}}\exp(s_{ij})} \\
        &+\log\frac{\exp(s_{ii})}{\sum_{j\in\mathcal{I}}\exp(s_{ji})}\bigg),
    \end{split}
    \end{equation}
    {where $s_{ij}=\exp(\gamma)\mathbf{a}_i^{\mathsf T}\mathbf{b}_j$ is the scaled similarity, and $\mathbf{a}_i$ and $\mathbf{b}_j$ are rows of $\mathbf{A}$ and $\mathbf{B}$. The learnable parameter $\gamma$ controls the positive logit scale $\exp(\gamma)$. The symmetric NCE loss increases the similarity between representations of the same user location and contrasts them against representations of other valid user locations in the mini-batch.}

    {For CLIP, $\bar{\mathbf{u}}_{\mathrm{Dec},k}$ and $\bar{\mathbf{u}}_{\mathrm{CSI},k}$ from the same user instance form a positive pair. Features from other valid user instances in the mini-batch provide the negative set. The CLIP alignment loss is}
    \begin{equation}
        \mathcal{L}_{\mathrm{Align}}^{\mathrm{CLIP}}
        =\lambda_{\mathrm{CLIP}}\mathcal{L}_{\mathrm{NCE}}(\bar{\mathbf{U}}_{\mathrm{Dec}},\bar{\mathbf{U}}_{\mathrm{CSI}}),
    \end{equation}
    {where $\lambda_{\mathrm{CLIP}}\geq0$ is the CLIP loss weight.}

    {For MISA, let $\mathbf{S}_\mu,\mathbf{C}_\mu\in\mathbb{R}^{|\mathcal{I}|\times d_{\mathrm{proj}}}$ stack the shared and private representations of all valid tokens from modality $\mu$. The similarity loss aligns the shared representations. The difference loss reduces overlap between the shared and private representations within each modality. The linear reconstruction head $g_{\mathrm{rec},\mu}$ maps their sum back to the projected-token space. These losses are}
    \begin{align}
        \mathcal{L}_{\mathrm{sim}}&=\mathcal{L}_{\mathrm{NCE}}(\bar{\mathbf{S}}_{\mathrm{Dec}},\bar{\mathbf{S}}_{\mathrm{CSI}}), \\
        \mathcal{L}_{\mathrm{diff}}&=\frac{1}{d_{\mathrm{proj}}^2}\sum_{\mu\in\mathcal{M}}\left\|(\mathbf{S}_{\mu}^{\mathrm{cn}})^{\mathsf T}\mathbf{C}_{\mu}^{\mathrm{cn}}\right\|_{F}^{2}, \\
        \mathcal{L}_{\mathrm{rec}}&=\frac{1}{2|\mathcal{I}|d_{\mathrm{proj}}}\sum_{\mu\in\mathcal{M}}\left\|g_{\mathrm{rec},\mu}(\mathbf{S}_{\mu}+\mathbf{C}_{\mu})-\mathbf{U}_{\mu}\right\|_{F}^{2}.
    \end{align}
    {where the superscript $(\cdot)^{\mathrm{cn}}$ denotes column centering followed by row-wise $\ell_2$ normalization. Since $(\mathbf{S}_{\mu}^{\mathrm{cn}})^{\mathsf T}\mathbf{C}_{\mu}^{\mathrm{cn}}$ has size $d_{\mathrm{proj}}\times d_{\mathrm{proj}}$, the factor $1/d_{\mathrm{proj}}^2$ {averages its squared entries for each modality}. The factor $1/(2|\mathcal{I}|d_{\mathrm{proj}})$ {averages the reconstruction error over the two modalities, all valid tokens, and all feature dimensions}. The reconstruction heads prevent the decomposition from discarding information in $\mathbf{U}_{\mu}$. Their outputs are used only in $\mathcal{L}_{\mathrm{rec}}$ and are not passed to the localization head. The MISA alignment loss is}
    \begin{equation}
        \mathcal{L}_{\mathrm{Align}}^{\mathrm{MISA}}
        =\lambda_{\mathrm{sim}}\mathcal{L}_{\mathrm{sim}}
        +\lambda_{\mathrm{diff}}\mathcal{L}_{\mathrm{diff}}
        +\lambda_{\mathrm{rec}}\mathcal{L}_{\mathrm{rec}}.
    \end{equation}
    {The nonnegative coefficients $\lambda_{\mathrm{sim}}$, $\lambda_{\mathrm{diff}}$, and $\lambda_{\mathrm{rec}}$ weight the similarity, difference, and reconstruction losses.}
    \endgroup

\experimentalsettingstable
\architecturetable

\newcommand{\overallresultsblock}{%
\begin{figure*}[t]

\centering
\makebox[\textwidth][c]{\includegraphics[width=0.64\textwidth]{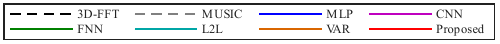}}\\[0.2mm]
\begin{subfigure}[t]{0.325\textwidth}
\centering
\includegraphics[width=\linewidth]{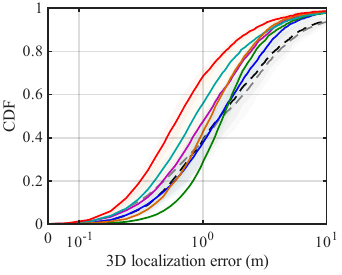}
\caption{DeepVerse 6G}
\end{subfigure}\hfill
\begin{subfigure}[t]{0.325\textwidth}
\centering
\includegraphics[width=\linewidth]{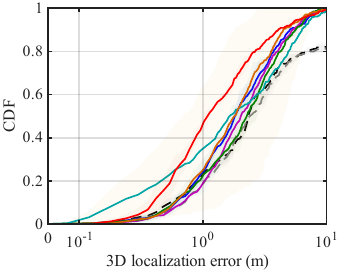}
\caption{Raymobtime}
\end{subfigure}\hfill
\begin{subfigure}[t]{0.325\textwidth}
\centering
\includegraphics[width=\linewidth]{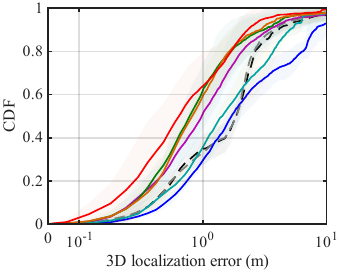}
\caption{Multimodal-Wireless}
\end{subfigure}
\caption{{The CDFs compare 3D localization errors on the test sets of the three datasets. Shaded regions indicate one standard deviation above and below the mean. }}
\label{fig:main_cdf_three_datasets}
\end{figure*}

\begin{table*}[t]

\centering
\caption{Overall 3D localization performance on all datasets.}
\label{tab:overall_three_datasets}
\scriptsize
\renewcommand{\arraystretch}{0.94}
\setlength{\tabcolsep}{3.0pt}
\begin{tabular*}{\textwidth}{@{\extracolsep{\fill}}llccccc}
\toprule
\textbf{Dataset} & \textbf{Method} & \textbf{Mean (m)} & \textbf{Median (m)} & \textbf{P95 (m)} & \textbf{Acc@1m (\%)} & \textbf{Acc@2m (\%)} \\
\midrule
\multirow{10}{*}{DeepVerse 6G}
& 3D-FFT & $3.93\pm1.05$ & $1.43\pm0.45$ & $13.84\pm7.59$ & $39.04\pm10.48$ & $61.58\pm9.20$ \\
& MUSIC & $4.15\pm1.17$ & $1.54\pm0.64$ & $14.26\pm6.64$ & $38.74\pm12.88$ & $58.29\pm10.84$ \\
& MLP & $2.20\pm0.43$ & $1.37\pm0.21$ & $6.15\pm1.33$ & $37.32\pm7.20$ & $64.59\pm6.06$ \\
& CNN & $1.77\pm0.15$ & $1.08\pm0.02$ & $5.03\pm0.64$ & $47.30\pm1.22$ & $73.63\pm0.82$ \\
& FNN & $2.19\pm0.36$ & $1.48\pm0.15$ & $5.89\pm1.14$ & $28.54\pm1.88$ & $68.38\pm5.01$ \\
& L2L & $1.73\pm0.49$ & $0.89\pm0.23$ & $5.48\pm2.10$ & $55.71\pm11.97$ & $79.78\pm7.35$ \\
& VAR & $1.78\pm0.12$ & $1.16\pm0.10$ & $4.84\pm0.77$ & $42.54\pm4.33$ & $74.42\pm1.63$ \\
& Proposed (None) & $1.31\pm0.12$ & $0.70\pm0.13$ & $3.93\pm0.74$ & $65.51\pm6.28$ & $85.83\pm0.61$ \\
& Proposed (CLIP) & $1.45\pm0.08$ & $0.80\pm0.13$ & $4.49\pm0.48$ & $59.94\pm5.79$ & $81.99\pm1.64$ \\
& \textbf{Proposed (MISA)} & $\mathbf{1.27\pm0.18}$ & $\mathbf{0.65\pm0.09}$ & $4.04\pm0.93$ & $\mathbf{68.25\pm4.16}$ & $\mathbf{86.21\pm1.80}$ \\
\midrule
\multirow{10}{*}{Raymobtime}
& 3D-FFT & $9.88\pm1.05$ & $2.37\pm0.09$ & $54.87\pm14.10$ & $22.62\pm2.30$ & $39.76\pm1.65$ \\
& MUSIC & $10.24\pm1.10$ & $2.45\pm0.32$ & $55.18\pm12.90$ & $19.05\pm1.80$ & $43.10\pm4.76$ \\
& MLP & $2.51\pm0.17$ & $1.90\pm0.22$ & $6.49\pm0.94$ & $23.33\pm2.51$ & $54.76\pm6.94$ \\
& CNN & $2.62\pm0.10$ & $2.04\pm0.30$ & $6.46\pm0.29$ & $18.57\pm3.78$ & $49.05\pm6.64$ \\
& FNN & $2.73\pm0.17$ & $2.29\pm0.41$ & $6.71\pm0.27$ & $22.38\pm1.49$ & $43.33\pm8.46$ \\
& L2L & $2.71\pm1.76$ & $2.25\pm2.18$ & $6.32\pm1.19$ & $34.76\pm28.31$ & $54.05\pm38.15$ \\
& VAR & $2.32\pm0.10$ & $1.81\pm0.21$ & $6.18\pm0.18$ & $24.76\pm5.27$ & $54.29\pm5.15$ \\
& \textbf{Proposed (None)} & $\mathbf{1.86\pm0.03}$ & $\mathbf{1.06\pm0.07}$ & $\mathbf{5.87\pm0.43}$ & $\mathbf{45.71\pm4.46}$ & $\mathbf{70.48\pm2.89}$ \\
& Proposed (CLIP) & $1.96\pm0.14$ & $1.33\pm0.19$ & $6.35\pm0.41$ & $35.48\pm7.15$ & $69.76\pm5.46$ \\
& Proposed (MISA) & $1.91\pm0.06$ & $1.21\pm0.07$ & $6.49\pm0.70$ & $41.43\pm1.24$ & $70.24\pm2.97$ \\
\midrule
\multirow{10}{*}{Multimodal-Wireless}
& 3D-FFT & $2.94\pm1.20$ & $1.83\pm0.19$ & $17.36\pm21.09$ & $34.61\pm7.10$ & $55.09\pm4.86$ \\
& MUSIC & $2.73\pm1.24$ & $1.85\pm0.13$ & $16.88\pm21.47$ & $32.96\pm5.18$ & $55.60\pm6.92$ \\
& MLP & $3.64\pm1.26$ & $1.78\pm0.32$ & $13.56\pm7.80$ & $29.48\pm3.07$ & $55.65\pm7.07$ \\
& CNN & $1.69\pm0.89$ & $1.02\pm0.30$ & $5.71\pm4.55$ & $51.01\pm11.96$ & $78.87\pm15.50$ \\
& FNN & $1.39\pm0.56$ & $0.79\pm0.04$ & $5.48\pm4.36$ & $61.72\pm6.34$ & $83.61\pm9.85$ \\
& L2L & $2.38\pm0.51$ & $1.48\pm0.18$ & $8.63\pm4.75$ & $35.47\pm4.15$ & $61.41\pm9.18$ \\
& VAR & $1.39\pm0.72$ & $0.87\pm0.29$ & $5.31\pm4.63$ & $58.84\pm15.23$ & $84.91\pm11.37$ \\
& \textbf{Proposed (None)} & $\mathbf{1.15\pm0.57}$ & $\mathbf{0.70\pm0.36}$ & $\mathbf{4.33\pm3.66}$ & $63.84\pm15.54$ & $\mathbf{86.02\pm9.66}$ \\
& Proposed (CLIP) & $1.32\pm0.46$ & $0.87\pm0.10$ & $4.80\pm3.93$ & $59.68\pm8.98$ & $85.12\pm8.56$ \\
& Proposed (MISA) & $1.36\pm0.56$ & $0.89\pm0.21$ & $4.87\pm4.23$ & $56.73\pm12.89$ & $84.76\pm10.51$ \\
\bottomrule
\end{tabular*}
\end{table*}
}

    \subsection{Localization Head}

    {Depending on the alignment choice, the shared localization head takes $\mathbf{q}_{k}^{\mathrm{None}}=[\mathbf{z}_{\mathrm{Dec},k};\mathbf{z}_{\mathrm{CSI},k}]$, $\mathbf{q}_{k}^{\mathrm{CLIP}}=[\bar{\mathbf{u}}_{\mathrm{Dec},k};\bar{\mathbf{u}}_{\mathrm{CSI},k}]$, or the four-part representation $\mathbf{q}_{k}^{\mathrm{MISA}}$ defined above, and passes the selected $\mathbf{q}_{k}$ to the same MLP localization head $f_{\mathrm{Loc}}$ to predict the 3D coordinates of user $k$:}
    \begin{equation}
        \hat{\mathbf{p}}_k=f_{\mathrm{Loc}}(\mathbf{q}_k)\in\mathbb{R}^{3}.
    \end{equation}
  
    \subsection{Training Objective and Strategy}

    Let $\hat{\mathbf{p}}_k$ and $\mathbf{p}_k$ denote the predicted and ground-truth 3D coordinates of the $k$-th valid user slot, respectively. Since the number of active users varies across samples, all training losses are computed only over valid user tokens indicated by the padding mask. The primary localization objective is defined as
    \begin{equation}
        \mathcal{L}_{\mathrm{Loc}}=\frac{1}{K}\sum_{k=1}^{K}\left\|\hat{\mathbf{p}}_k-\mathbf{p}_k\right\|_2,
    \end{equation}
    where $K$ denotes the number of valid user tokens.
    \begingroup
    
    {All variants use the same total objective}
    \begin{equation}
        \mathcal{L}_{\mathrm{Total}}
        =\mathcal{L}_{\mathrm{Loc}}+\mathcal{L}_{\mathrm{Align}}.
        \label{eq:total_loss}
    \end{equation}
    {The alignment term is set to $\mathcal{L}_{\mathrm{Align}}^{\mathrm{CLIP}}$ for CLIP, $\mathcal{L}_{\mathrm{Align}}^{\mathrm{MISA}}$ for MISA, and zero without alignment.}
    
    {Training proceeds in two stages. The CSI encoder is first trained with a CSI-only localization head. Its parameters then initialize the multimodal model and remain frozen while the visual projection, Transformer decoder, alignment module, and localization head are optimized.}
    \endgroup

\begingroup

\captionsetup[figure]{font=footnotesize}
\captionsetup[table]{font=footnotesize}
\section{Experimental Results}
\label{IV}

\subsection{Experimental Setup}

\subsubsection{Datasets and Data Selection}
\label{sec:datasets_data_selection}
We evaluate on three multimodal datasets: DeepVerse 6G, Raymobtime, and Multimodal-Wireless \cite{DeepVerse, Raymobtime, Multimodal-Wireless}. DeepVerse 6G serves as the primary benchmark for evaluation {across \textbf{multiple BSs}} and for the main architectural analyses. Raymobtime contains \textbf{wireless LoS and NLoS} labels, allowing us to compare localization performance under the two propagation conditions. Multimodal-Wireless provides matched sunny, rainy, and foggy recordings along the same trajectories, allowing us to compare localization performance under \textbf{different visual weather conditions}. Following the task definition in Section~\ref{II}, we select frames containing at least one communicating user inside the camera FoV and configured forward range. 

The O1 scenario of DeepVerse 6G contains {BSs 1--4. BS sites 2, 3, and 4} serve as the test set in three separate experiments. For each experiment, {the other three BSs} are split 9:1 into training and validation sets. {At each site, scenes are ordered by scene index, with the first 90\% used for training and the rest for validation.} Raymobtime combines the \texttt{s008} and \texttt{s009} scenarios, which model {mobile receivers in the Rosslyn environment} \cite{Raymobtime}. The data are split 70:15:15 into training, validation, and test sets. Multimodal-Wireless combines the {five way intersection scenario} in Town03 and the {wide skybridge scenario} in Town10 \cite{Multimodal-Wireless}. The data use the same 70:15:15 split. {Within each scenario, the split is made using non-overlapping samples of 100 frames. Fig.~\ref{fig:dataset_split_counts} shows the number of samples for each value of the number of users in each split. Each sample belongs to only one split, although neighboring samples may belong to different splits. Models are trained and tested separately on each dataset.}

\begin{figure*}[t]
\centering
\begin{subfigure}[t]{0.325\textwidth}
\centering
\includegraphics[width=\linewidth]{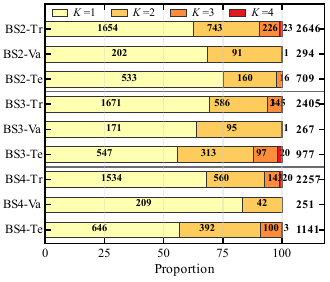}
\caption{{DeepVerse 6G}}
\label{fig:dataset_split_counts_deepverse}
\end{subfigure}\hfill
\begin{subfigure}[t]{0.325\textwidth}
\centering
\includegraphics[width=\linewidth]{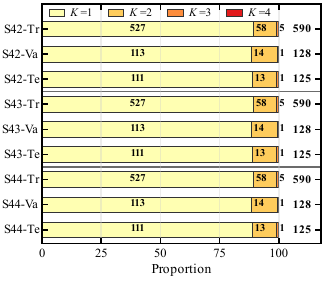}
\caption{{Raymobtime}}
\label{fig:dataset_split_counts_raymobtime}
\end{subfigure}\hfill
\begin{subfigure}[t]{0.325\textwidth}
\centering
\includegraphics[width=\linewidth]{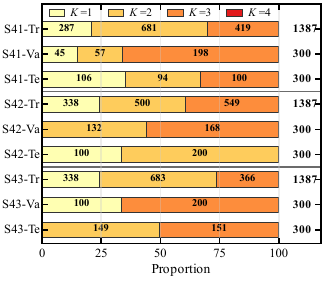}
\caption{{Multimodal-Wireless}}
\label{fig:dataset_split_counts_multimodal_wireless}
\end{subfigure}
\caption{{Training, validation, and test splits and their number of users ($K$) distributions. Numbers denote the number of samples.}}
\label{fig:dataset_split_counts}
\end{figure*}

{We adapt the channel data from Raymobtime and Multimodal-Wireless to use the same CSI input format as DeepVerse 6G.} Raymobtime dataset provides ray-tracing parameters for each propagation path rather than array CSI that can be used directly. We select {up to $P_k$ strongest available paths} and construct the complex CSI from their powers, delays, departure angles, and a fixed realization of the path phases {sampled from $\mathcal{U}[0,2\pi)$}. The array response is generated for a half-wavelength $4\times4$ UPA. Multimodal-Wireless dataset instead provides channel parameters for 16-element uniform linear arrays (ULAs). For each path, we use the complex gain at the reference antenna pair together with the provided departure angles and delay to reconstruct the response of the same half-wavelength $4\times4$ UPA. Both adaptations produce CSI that is consistent with the input format used for DeepVerse 6G dataset. {When fewer than $P_k$ valid paths are available, Raymobtime pads unused ray records with NaNs, whereas Multimodal-Wireless zero-pads unused path entries and stores the actual valid-path count. These padded entries are excluded during CSI synthesis and are therefore neither duplicated nor assigned identical path parameters.} {Table~\ref{tab:experimental_settings} summarizes the dataset settings and common implementation parameters.}

\newcommand{\encoderablationtable}{%
\begin{table*}[t]

\centering
\caption{Localization performance with different CSI encoders on all datasets.}
\label{tab:encoder_ablation}
\scriptsize
\setlength{\tabcolsep}{3.5pt}
\begin{tabular*}{\textwidth}{@{\extracolsep{\fill}}llccccc}
\toprule
\textbf{Dataset} & \textbf{CSI encoder} & \textbf{Mean (m)} & \textbf{Median (m)} & \textbf{P95 (m)} & \textbf{Acc@1m (\%)} & \textbf{Acc@2m (\%)} \\
\midrule
\multirow{4}{*}{DeepVerse 6G}
& MLP & $1.29\pm0.14$ & $0.63\pm0.07$ & $4.65\pm1.08$ & $67.80\pm3.21$ & $84.41\pm2.14$ \\
& \textbf{CNN} & $\mathbf{1.27\pm0.18}$ & $0.65\pm0.09$ & $\mathbf{4.04\pm0.93}$ & $\mathbf{68.25\pm4.16}$ & $\mathbf{86.21\pm1.80}$ \\
& FNN & $1.36\pm0.08$ & $0.77\pm0.18$ & $4.35\pm0.66$ & $61.04\pm9.70$ & $82.87\pm3.62$ \\
& L2L & $1.44\pm0.30$ & $0.73\pm0.10$ & $4.34\pm1.41$ & $64.23\pm5.14$ & $85.10\pm4.66$ \\
\midrule
\multirow{4}{*}{Raymobtime}
& MLP & $1.88\pm0.19$ & $1.07\pm0.15$ & $6.39\pm0.80$ & $\mathbf{48.10\pm5.46}$ & $70.24\pm6.16$ \\
& CNN & $2.02\pm0.16$ & $1.35\pm0.15$ & $6.37\pm0.10$ & $38.33\pm3.22$ & $67.38\pm5.36$ \\
& \textbf{FNN} & $\mathbf{1.86\pm0.03}$ & $\mathbf{1.06\pm0.07}$ & $\mathbf{5.87\pm0.43}$ & $45.71\pm4.46$ & $\mathbf{70.48\pm2.89}$ \\
& L2L & $2.49\pm1.64$ & $1.79\pm1.78$ & $7.08\pm1.65$ & $40.48\pm32.20$ & $61.19\pm33.29$ \\
\midrule
\multirow{4}{*}{Multimodal-Wireless}
& MLP & $3.65\pm1.24$ & $1.73\pm0.24$ & $13.50\pm7.28$ & $24.34\pm7.88$ & $56.61\pm6.03$ \\
& CNN & $1.59\pm0.76$ & $0.88\pm0.17$ & $5.65\pm3.94$ & $56.98\pm8.83$ & $79.43\pm13.25$ \\
& \textbf{FNN} & $\mathbf{1.15\pm0.57}$ & $\mathbf{0.70\pm0.36}$ & $\mathbf{4.33\pm3.66}$ & $\mathbf{63.84\pm15.54}$ & $\mathbf{86.02\pm9.66}$ \\
& L2L & $2.21\pm0.41$ & $1.42\pm0.11$ & $7.84\pm4.85$ & $36.10\pm5.26$ & $63.26\pm4.90$ \\
\bottomrule
\end{tabular*}
\end{table*}
}

\newcommand{\robustnessfigure}{%
\begin{figure*}[t]

\centering
\makebox[\textwidth][c]{\includegraphics[width=0.76\textwidth]{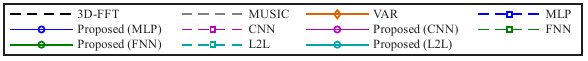}}\\[0.2mm]
\begin{subfigure}[t]{0.325\textwidth}
\centering
\includegraphics[width=\linewidth]{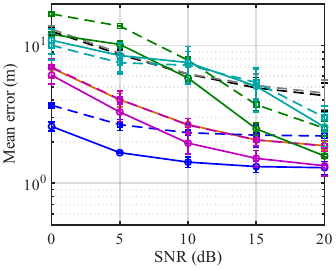}
\caption{DeepVerse 6G: SNR}
\end{subfigure}\hfill
\begin{subfigure}[t]{0.325\textwidth}
\centering
\includegraphics[width=\linewidth]{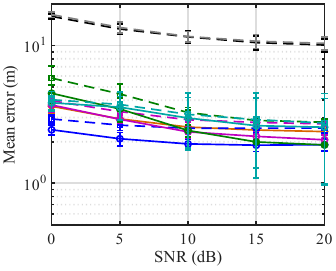}
\caption{Raymobtime: SNR}
\end{subfigure}\hfill
\begin{subfigure}[t]{0.325\textwidth}
\centering
\includegraphics[width=\linewidth]{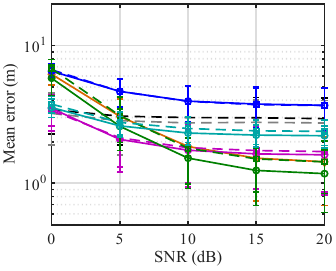}
\caption{Multimodal-Wireless: SNR}
\end{subfigure}\\[1.0mm]
\begin{subfigure}[t]{0.325\textwidth}
\centering
\includegraphics[width=\linewidth]{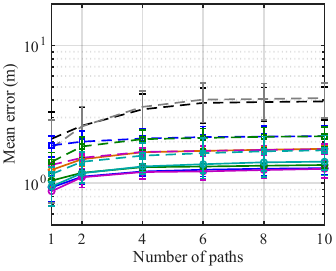}
\caption{DeepVerse 6G: number of paths}
\end{subfigure}\hfill
\begin{subfigure}[t]{0.325\textwidth}
\centering
\includegraphics[width=\linewidth]{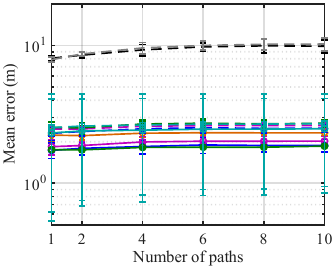}
\caption{Raymobtime: number of paths}
\end{subfigure}\hfill
\begin{subfigure}[t]{0.325\textwidth}
\centering
\includegraphics[width=\linewidth]{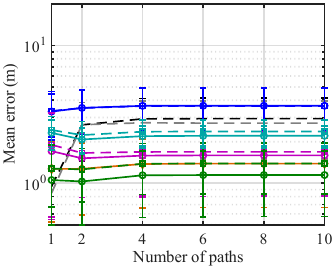}
\caption{Multimodal-Wireless: number of paths}
\end{subfigure}
\caption{{Localization performance under different SNRs and numbers of paths on all datasets.}}
\label{fig:robustness_sweeps}
\end{figure*}
}

\overallresultsblock
\encoderablationtable
\robustnessfigure

\subsubsection{Baselines}
The baselines comprise traditional model-based methods, deep learning CSI methods, and a multimodal fusion method.
\begin{itemize}
    \setlength{\itemsep}{0pt}
    \setlength{\parskip}{0pt}
    \setlength{\parsep}{0pt}
    \setlength{\topsep}{2pt}
    \item \textbf{3D-FFT} extracts dominant angle-delay peaks from the CSI.
    \item \textbf{MUSIC} estimates angles and delays from CSI subspaces.
    \item \textbf{Multilayer perceptron (MLP)} processes flattened real and imaginary CSI components.
    \item \textbf{Convolutional neural network (CNN)} learns patterns on the antenna-subcarrier plane with two-dimensional convolutions.
    \item \textbf{Factorized neural network (FNN)} applies separate convolutional operations to the antenna and frequency axes.
    \item \textbf{Learn to Localize (L2L)} adapts the three-dimensional CNN to our CSI input and coordinate target \cite{learn-to-localize}.
    \item \textbf{Vision-aided Radio (VAR)} combines each CSI token with a globally pooled visual feature \cite{de2020vision}.
\end{itemize}

{Unless configurations are compared separately, Proposed denotes the best combination of CSI encoder and feature-processing method for each dataset, selected from Tables~~\ref{tab:overall_three_datasets} and \ref{tab:encoder_ablation}.} {Table~\ref{tab:model_architectures} summarizes the configurations of all compared methods.}

\subsubsection{Evaluation Metrics}
We report five localization metrics: mean error, median error, P95, $\mathrm{Acc}@1\mathrm{m}$, and $\mathrm{Acc}@2\mathrm{m}$. The first three metrics are measured in meters, while the two accuracy metrics are reported as percentages. Index the $N$ valid user instances in the test set by $i\in\{1,\ldots,N\}$. For the ground-truth position $\mathbf{p}_i$ and estimated position $\hat{\mathbf{p}}_i$ of user instance $i$, the 3D localization error is
\begin{equation}
e_i=\lVert\hat{\mathbf{p}}_i-\mathbf{p}_i\rVert_2.
\end{equation}

The mean error is $N^{-1}\sum_{i=1}^{N}e_i$. The median error is the 50th percentile of $\{e_i\}_{i=1}^{N}$, and P95 is its 95th percentile. For a distance threshold $\epsilon>0$, the localization accuracy is
\begin{equation}
\mathrm{Acc}@\epsilon =\frac{100}{N}\sum_{i=1}^{N}\mathbb{I}(e_i\le \epsilon)\;[\%],
\end{equation}
where $\mathbb{I}(e_i\le \epsilon)$ equals one when the localization error does not exceed $\epsilon$ and zero otherwise. We obtain $\mathrm{Acc}@1\mathrm{m}$ and $\mathrm{Acc}@2\mathrm{m}$ by setting $\epsilon$ to 1~m and 2~m, respectively.

{For DeepVerse 6G, we average results across the three BS-site experiments described in Sec.~\ref{sec:datasets_data_selection}. For Raymobtime and Multimodal-Wireless, we use three random data splits for each dataset. Each metric is computed over all valid UEs in each test set and reported as the mean $\pm$ standard deviation across the three experiments. }

To evaluate the effect of channel estimation errors on localization performance, we add additive white Gaussian noise (AWGN) to the CSI only during inference at SNRs of $\{0,5,10,15,20\}$~dB. To evaluate the effect of the {maximum number of included paths}, the models are trained using CSI constructed from {up to ten strongest available paths}. During inference, the CSI is reconstructed using limits of $P\in\{1,2,4,6,8,10\}$. If sample $i$ contains {$N_i$ valid paths}, its CSI includes the {strongest $\min(P,N_i)$ paths}. The ray-tracing path records are used only for {CSI construction} and are not inputs to the localization models.

\subsection{Overall Localization Performance}

Fig.~\ref{fig:main_cdf_three_datasets} illustrates the localization error cumulative distribution functions (CDFs) of all methods on the three datasets. Table~\ref{tab:overall_three_datasets} reports their typical statistics.
The {Proposed} model shows a consistent advantage across all three datasets. It shifts the central part of the error distribution to the left, while the gains at $\mathrm{Acc}@1\mathrm{m}$ and $\mathrm{Acc}@2\mathrm{m}$ show that more users are localized within small error ranges. The advantage becomes more pronounced on the Raymobtime dataset, where the model-based estimators develop much larger errors and heavier tails. This is consistent with the more challenging propagation conditions in this dataset, since \texttt{s009} contains much higher ratios of NLoS users than the other two datasets \cite{Raymobtime}, while the {Proposed} model remains relatively stable. {MISA} performs best on the DeepVerse 6G dataset, while {the version without alignment} gives the strongest overall result on the Multimodal-Wireless and Raymobtime datasets. All three alignment methods for the Raymobtime dataset use {FNN}. {MISA} remains closer to {the version without alignment} than {CLIP} in the mean, median, and threshold accuracies. The improvement is also not limited to the center of the distribution. The {Proposed} model generally keeps more samples in the low error region while reducing large localization errors, showing improvements in both localization precision and reliability. Finally, the gain remains visible even on the Multimodal-Wireless dataset, where several learning-based baselines already perform well.

\subsection{Ablation Experiments}
\begin{figure}[t]
\centering
\includegraphics[width=\columnwidth]{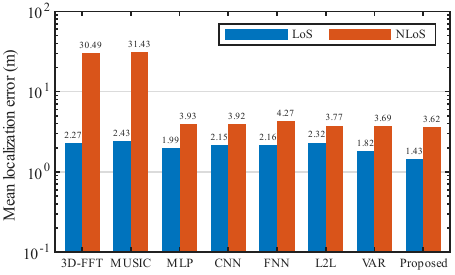}
\caption{{Comparison of mean localization performance under wireless LoS and NLoS conditions on the Raymobtime dataset.}}
\label{fig:raymobtime_los_nlos}
\end{figure}
\subsubsection{CSI Encoder Selection}

Table~\ref{tab:encoder_ablation} compares the four CSI encoders within the {Proposed} model while holding the alignment method fixed for each dataset.
The choice of CSI encoder is clearly dataset-dependent. {CNN} gives the strongest overall performance on the DeepVerse 6G dataset, while {FNN} performs best on the Raymobtime and Multimodal-Wireless datasets, showing that no single encoder is optimal across all environments. {FNN} is particularly effective on the more challenging datasets. On the Raymobtime and Multimodal-Wireless datasets, it improves both the central part of the error distribution and the tail behavior, while also giving stronger results at $\mathrm{Acc}@1\mathrm{m}$ and $\mathrm{Acc}@2\mathrm{m}$. In contrast, {MLP} and {L2L} are generally weaker, and their gap becomes more visible on the harder datasets. This suggests that {CNN} and {FNN} are better able to extract CSI features that remain useful under more complex propagation conditions.

\subsubsection{Visual Dependence}

Table~\ref{tab:visual_ablation} compares the mean localization error under normal visual input and four visual input perturbations. 
The global-shuffle condition replaces each test image with a different image through a fixed derangement while keeping the CSI observations and ground-truth positions unchanged. In the RGB-features-zero condition, the visual feature map is set to zero while retaining the visual positional encoding. In the visual-position-zero condition, the positional encoding is set to zero while retaining the RGB feature map. In the visual-memory-zero condition, both components of the visual memory are set to zero. All four interventions are applied only at inference using the same trained model and test samples as the normal condition. All visual input perturbations lead to higher localization error across the three datasets, confirming that the Proposed model makes effective use of visual information rather than relying only on CSI. In particular, global shuffling causes a clear degradation on the DeepVerse 6G and Raymobtime datasets, showing that scene-consistent visual context is important. The contribution of each visual component is also dataset-dependent. The DeepVerse 6G dataset is more sensitive to RGB features, while the Multimodal-Wireless dataset shows a larger dependence on visual position and visual memory, suggesting that the Proposed model uses different visual cues under different environments.

\subsection{\texorpdfstring{{Performance under Different Test Conditions}}{Performance under Different Test Conditions}}

\subsubsection{Different SNRs and Numbers of Paths}

\begin{table}[t]
\centering
\caption{Visual-input ablation performance on all datasets.}
\label{tab:visual_ablation}
\scriptsize
\setlength{\tabcolsep}{2.6pt}
\begin{tabular*}{\columnwidth}{@{\extracolsep{\fill}}lccc}
\toprule
\textbf{Condition} & \textbf{DeepVerse 6G} & \textbf{Raymobtime} & \textbf{MM-Wireless} \\
\midrule
Normal & $\mathbf{1.27\pm0.18}$ & {$\mathbf{1.86\pm0.03}$} & $\mathbf{1.15\pm0.57}$ \\
Global shuffle & $2.37\pm0.30$ & {$3.32\pm0.31$} & $1.32\pm0.48$ \\
RGB features zero & $2.87\pm0.25$ & {$3.06\pm0.30$} & $1.34\pm0.47$ \\
Visual position zero & $2.34\pm0.31$ & {$2.75\pm0.09$} & $1.50\pm0.30$ \\
Visual memory zero & $1.99\pm0.15$ & {$3.08\pm0.10$} & $1.69\pm0.13$ \\
\bottomrule
\end{tabular*}
\end{table}

\begin{table}[t]
\centering
\caption{Localization performance under different numbers of active users on all datasets.}
\label{tab:user_count}
\scriptsize
\setlength{\tabcolsep}{2.8pt}
\begin{tabular*}{\columnwidth}{@{\extracolsep{\fill}}lcccc}
\toprule
\textbf{Dataset} & \textbf{$K=1$} & \textbf{$K=2$} & \textbf{$K=3$} & \textbf{$K=4$} \\
\midrule
DeepVerse 6G & $0.92\pm0.17$ & $1.53\pm0.43$ & $1.77\pm0.48$ & $1.46\pm0.31^{\dagger}$ \\
Raymobtime & $1.99\pm0.10$ & $2.21\pm0.32$ & $1.54\pm0.88^{\dagger}$ & -- \\
Multimodal-Wireless & $1.21\pm0.30$ & $1.06\pm0.56$ & $1.55\pm0.90$ & -- \\
\bottomrule
\end{tabular*}
\vspace{0.4mm}
\parbox{\columnwidth}{\raggedright\scriptsize\textit{Note:} $\dagger$ denotes a condition with a limited number of test samples.}
\end{table}

\begin{figure}[t]
\centering
\includegraphics[width=\columnwidth]{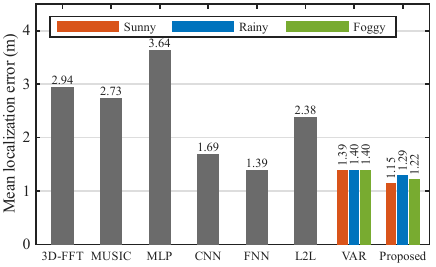}
\caption{{Comparison of mean localization performance under different weather conditions.}}
\label{fig:multimodal_wireless_weather}
\end{figure}

Fig.~\ref{fig:robustness_sweeps} illustrates the localization performance under different SNRs and {maximum numbers of included paths}. Localization performance generally improves as SNR increases, with the largest gains appearing from low to moderate SNRs. At higher SNRs, most learning-based methods gradually reach a stable error level, suggesting that noise is no longer the main limiting factor. The {Proposed} model also maintains a clear advantage under low SNR conditions, especially on the Raymobtime dataset, where the model-based estimators remain much less accurate even as SNR increases. This shows that improving signal quality alone cannot fully compensate for the difficulty of complex propagation environments. In comparison, the {maximum number of included paths} has a weaker effect on localization performance. Most learning-based methods remain relatively stable as this limit increases, while some model-based estimators are more sensitive to additional paths. Finally, a clear error floor remains at high SNR, particularly on the Raymobtime dataset.

\subsubsection{Different Numbers of Active Users}

Table~\ref{tab:user_count} reports the mean localization error for different numbers of active users. {Entries marked $\dagger$ contain few test samples, limiting comparisons across $K$.} {The error does not always increase with the number of users. It decreases from $K=2$ to $K=3$ on Raymobtime and from $K=1$ to $K=2$ on Multimodal-Wireless.}

\subsubsection{Wireless LoS and NLoS Conditions}

In the Raymobtime dataset, a wireless link is labeled LoS when its ray record {contains a direct path} and NLoS otherwise. This label describes {wireless propagation rather than visual visibility}. {Fig.~\ref{fig:raymobtime_los_nlos} compares the mean localization errors of all methods under wireless LoS and NLoS conditions. Every method has a higher mean error under NLoS propagation. The increase is especially large for 3D-FFT and MUSIC, while the learning-based methods show substantially smaller increases. This is because 3D-FFT and MUSIC treat the dominant path as a direct LoS path when converting angle and delay estimates into a position, which introduces bias under NLoS propagation. Proposed achieves the lowest mean error among the compared methods under both conditions.}

\begin{table}[t]

\centering
\caption{Computational complexity on the DeepVerse 6G dataset.}
\label{tab:complexity}
\scriptsize
\setlength{\tabcolsep}{1.6pt}
\begin{tabular*}{\columnwidth}{@{\extracolsep{\fill}}lcccc}
\toprule
\textbf{Method} & \textbf{Params (M)} & \makecell{\textbf{GFLOPs/}\\\textbf{sample}} & \makecell{\textbf{Train/Inf.}\\\textbf{Time (s/ms)}} & \makecell{\textbf{Train/Inf.}\\\textbf{Mem. (MB)}} \\
\midrule
3D-FFT & 0 & -- & --/1.81 & --/6.5 \\
MUSIC & 0 & -- & --/5.91 & --/3.6 \\
MLP & 0.544 & 0.0011 & 0.04/0.68 & 75.3/68.2 \\
CNN & 0.242 & 0.024 & 0.05/1.06 & 183.8/67.9 \\
FNN & 0.063 & 0.0036 & 0.06/0.61 & 137.6/64.6 \\
L2L & 1.169 & 0.065 & 0.24/6.91 & 182.6/74.5 \\
VAR & 3.164 & 7.435 & 10.42/4.12 & 1018.9/106.2 \\
Proposed (None) & 6.327 & 7.438 & 12.65/10.31 & 1148.3/130.0 \\
Proposed (CLIP) & 6.590 & 7.438 & 12.52/15.94 & 1150.7/132.0 \\
Proposed (MISA) & 6.476 & 7.438 & 13.41/15.76 & 1149.0/131.1 \\
\bottomrule
\end{tabular*}
\end{table}

\subsubsection{Different Weather Conditions}

Fig.~\ref{fig:multimodal_wireless_weather} compares localization performance on the Multimodal-Wireless dataset under sunny, rainy, and foggy visual conditions. All models are trained on sunny data. The rainy and foggy recordings follow the same Town 03 and Town 10 trajectories. At test time, we replace each sunny RGB frame with its synchronized weather counterpart while keeping the CSI and ground-truth position unchanged. This experiment isolates visual weather effects by keeping CSI fixed. Overall, the localization performance remains relatively stable across sunny, rainy, and foggy conditions. The Proposed model shows similar mean error levels across the three weather settings.

\subsection{Complexity Analysis}

Table~\ref{tab:complexity} reports the parameter count, FLOPs per sample, training and inference time, and training and inference memory under the DeepVerse 6G implementation protocol. Values before and after the slash give the training and inference measurements, respectively. The learned-model timing and memory measurements are obtained on an NVIDIA GeForce RTX 5070 Laptop GPU. {MUSIC} is timed on the CPU because its eigendecomposition implementation is CPU based.

The multimodal methods require substantially more computation than the CSI-only baselines, with most of the cost coming from the visual backbone. {VAR} and the {Proposed} variants have nearly identical FLOPs, indicating that the additional cross-modal interaction contributes little to the total arithmetic cost. Compared with {VAR}, the proposed model mainly increases the parameter count and inference latency because of the Transformer-based cross-attention. Overall, the proposed model trades a moderate increase in complexity for the stronger localization performance observed in the previous experiments.

\endgroup

\section{Conclusion}
\label{V}

    {{This paper presented a vision--wireless fusion framework for multi-user localization using pilot-indexed CSI. Pilot-indexed CSI preserves communicating-user identity, while spatial visual memory provides scene context. Cross-attention uses each CSI token to retrieve visual information for the corresponding UE before localization.} Experiments on three datasets consistently show that the proposed model improves localization over CSI-only and global-fusion baselines, with clearer gains {in the evaluated low SNR and NLoS cases}. The ablation studies further confirm that the model makes effective use of visual information and that correct {vision--wireless fusion} is important. At the same time, the results also show that there is no single fixed combination of CSI encoder and alignment method that performs best across all datasets, suggesting that the best multimodal representation depends on the propagation and scene characteristics.}

    {Future work can extend the proposed framework in several directions. First, multi-view camera information can be integrated to provide broader visual coverage and reduce the dependence on a single camera FoV. This may also improve localization when users are occluded or only partially visible from one viewpoint. Second, a more general cross-scene localization framework can be developed to improve generalization to unseen sites, propagation environments, and carrier frequencies, reducing the need to retrain the model for every new deployment scenario. Finally, lightweight visual encoders and more efficient cross-modal interaction can be explored to reduce computational cost and support denser multi-user scenarios.
}


\balance
\bibliographystyle{IEEEtran}
\bibliography{IEEEabrv, ref}
\end{document}